\documentclass{article}

\usepackage[final,nonatbib]{nips_2017}

\usepackage[utf8]{inputenc} 
\usepackage[T1]{fontenc}    
\usepackage{hyperref}       
\usepackage{url}            
\usepackage{booktabs}       
\usepackage{amsfonts}       
\usepackage{nicefrac}       
\usepackage{microtype}      
  \usepackage{amsmath,amssymb,mathtools}
  \usepackage{color}
  \usepackage{enumitem}
  \usepackage{subcaption}
  \def\thetitle{Large Neural Network Based Detection of \\ Apnea, Bradycardia and Desaturation Events}
\title{\thetitle}

\def\ki{Dept. Women’s and Children’s Health, Karolinska Institutet, and Pediatric Emergency, Karolinska University Hospital, SE-171 76 Solna, Sweden. 
}
\def\kth{Dept. Information Science and Engineering, KTH Royal Institute of Technology, Stockholm, Sweden.}
\author{
	Antoine Honoré\thanks{\kth}~~\thanks{\ki}
	\hspace {1cm} Veronica Siljehav\footnotemark[2] \hspace {1cm} 
	Saikat Chatterjee\footnotemark[1] \hspace {1cm} 
	Eric Herlenius\footnotemark[2] 
	\\
	\texttt{honore@kth.se} \hspace {0.2cm}  \texttt{veronica.siljehav@ki.se} \, \hspace {0.2cm}  \texttt{sach@kth.se} \, \hspace {0.2cm}  \texttt{eric.herlenius@ki.se}
}

\begin{document}

\maketitle

\vspace{-0.5cm}
\begin{abstract}
Apnea, bradycardia and desaturation (ABD) events often precede life-threatening events including sepsis in newborn babies. Here, we explore machine learning for detection of ABD events as a binary classification problem. We investigate the use of a large neural network to achieve a good detection performance. To be user friendly, the chosen neural network does not require a high level of parameter tuning. Furthermore, a limited amount of training data is available and the training dataset is unbalanced. Comparing with two widely used state-of-the-art machine learning algorithms, the large neural network is found to be efficient. Even with a limited and unbalanced training data, the large neural network provides a detection performance level that is feasible to use in clinical care.
\end{abstract}

\vspace{-0.5cm}
\section{Introduction}
\vspace{-0.1cm}
Cardio-respiratory dysfunction and sepsis are major causes of morbidity and mortality in the neonatal population \cite{Liu2012}. Heart rate characteristic (HRC) monitoring system helps to reduce mortality of preterm babies \cite{Fairchild2013} and such systems have some success in detecting irregularities before they are clinically prominent \cite{moorman2011mortality,Fairchild_2010,lake2014complex,Sullivan2015,Moss2016}.
Apnea – breathing pauses and hypoventilation; Bradycardia – heart frequency reduction and Desaturation (ABD) events occur in daily life especially in preterm infants \cite{Hofstetter2008}. However, increasing ABD events also precede and may predict infection and life threatening events \cite{Siljehav2015}. Clinical symptoms of ABD events are: (a) apnea – breathing pauses and irregular breathing amplitudes; (b) bradycardia – heart frequency goes lower than usual levels; (c) desaturation – blood oxygen saturation is below normal levels; please see \cite{Lee2012} for more details. Depending on patient health and age, ABD events may occur in varying frequency. A high false negative rate in detecting ABD events is detrimental to patient care and a high false positive rate results in exhausting medical resources. Accurate detection of ABD events is a difficult task.  Naturally we are interested to use advanced machine learning for the detection task. Use of advanced machine learning methods by medical practitioners is often challenging. Many machine learning methods, like deep neural networks \cite{Hinton_DeepLearning_NaturePaper_2015,Hinton_Deng_2012} typically require a large amount of training data. However, even with a high level of collaboration with hospitals, we so far are unable to access a large amount of training data for ABD event detection task. Further, advanced machine learning methods often require technical knowledge on tuning their parameters for achieving a good performance. Medical practitioners typically prefer smart engineering tools without high involvement in technical parts. 
 
Recently a large neural network is proposed in \cite{Chatterjee2017}. This is called progressive learning network (PLN) where the network structure is determined in an autonomous way. The PLN learns its architecture in a data driven manner. At the time of training, the PLN automatically learns the required number of layers in its architecture and the number of nodes in its architecture. A learned PLN is expected to be large in the sense of its wideness and deepness. Training of the PLN is user friendly as it does not require pain-stalking involvements for tuning its parameters. Software codes are also available online. In \cite{Chatterjee2017} it was experimentally shown that PLN can provide good pattern recognition accuracy for several object recognition tasks, mainly image based objects. Here, we explore PLN for ABD event detection. An important point is that PLN was not shown to be tested on a limited training data that we explore in the present study. Further, the training dataset we have access to is unbalanced. Here unbalanced means that the training dataset has a very limited amount of labeled data with true ABD events compared to the amount of labeled data with no ABD events. PLN was not shown to be tested on unbalanced dataset. Hence, from a machine learning perspective, it is interesting to see how the PLN performs for limited as well as unbalanced dataset. We expect that PLN will provide a good performance for such data as the PLN learning process is highly regularized. At this point, we found in the literature one ABD event detection algorithm \cite{Lee2012} based on classical signal processing methods (filtering and thresholding). In practice, a thresholding based technique requires a high level of tuning, even at individual patient level. We are more interested in smart machine learning methods in this study, and refrain to pursue the existing thresholding based method. 
In this article, along with PLN we perform a comparative study with other machine learning algorithms. For this study, we investigate support-vector-machine (SVM) \cite{SVM_1999} and cross-entropy minimization based artificial neural network (ANN) \cite{bishop2006}. The choice of the two algorithms -- SVM and ANN -- is motivated by their popularity and success in various machine learning applications. Our study was performed in accordance with governmental ethical guidelines. The regional ethics committees of the hospital and municipality approved the research study.

\section{Cohort}
\begin{figure}[!ht]
\begin{minipage}[l]{0.6\textwidth}
\center
\includegraphics[width=\textwidth]{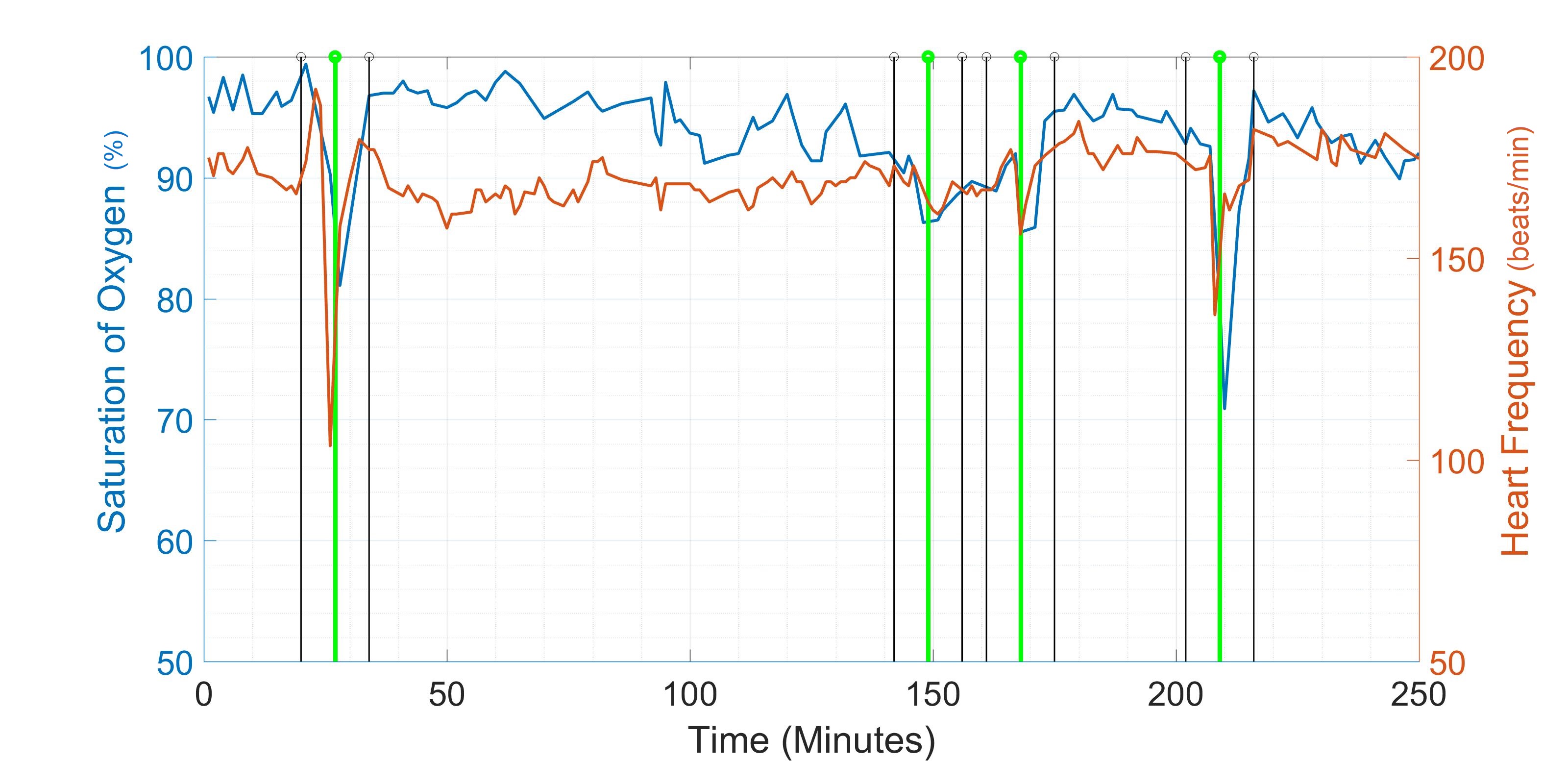}
\end{minipage}
\begin{minipage}[r]{0.37\textwidth}
\caption{~\\250 minutes of signal, showing four manually labeled ABD events region in a black box. The green line corresponds to the manually labeled point where an ABD event occurred. }
\label{fig:blackbox}
\end{minipage}
\end{figure}

\vspace{-0.2cm}
The clinical records of 13 eligible pre-mature infants were harvested at the neonatal intensive care units (ICU), Karolinska hospital, Stockholm, Sweden. The age of the babies at discharge was $ 28.57 \pm 23.20$ days and the total amount of acquired signals was 371.35 days.
The database is called Clinisoft\textsuperscript{\textregistered} and contains patient data collected at the ICU including diagnosis codes and notes, typically manually entered by clinicians. The signals we had access to are the heart frequency signal, taken over a minute, and the oxygen saturation level in the blood, averaged from a one minute time frame of a 100 Hz signal. Therefore both are sampled at the frequency of $\frac{1}{60}Hz$. For clarity and mathematical development later in the article, we denote the heart frequency signal by $x_1$ and the oxygen saturation level signal by $x_2$. Like many other natural signals -- such as speech, audio, image, electro-cardiograms, etc. -- we assume that the $x_1$ and $x_2$ signals are non-stationary. That means statistical property of a signal varies with respect to time. To handle the non-stationary aspect, we use frame-by-frame approach which is standard in speech, audio and image processing. We assume that an $l$-sample frame (or window) is statistically stationary, where the value of $l$ is not large. To collect feature vectors, we use a window shift by one sample. That means an $l$-sample window is moved across each of signals $x_1$ and $x_2$ by one sample in time-synchronization. This one-sample-shift helps to generate a large number of feature vectors from the available amount of signals. For a windowed portion of the signals, we collect all samples of the two signals $x_1$ and $x_2$, and construct a feature vector. Let us denote a feature vector by $\mathbf{x}$. The feature vector $\mathbf{x}$ comprises of $l$ consecutive samples of $x_1$ and $x_2$. That means $\mathbf{x}$ is of $2l$ dimension. We formally write $\mathbf{x} \in \mathbb{R}^D$ where $D=2l$. We use $l=15$. That means, we use 30-dimensional feature vector $\mathbf{x}$ in our study. In total we have $672 + 533866 = 534538$ feature vector instances, with following class labels.
\begin{enumerate}[noitemsep,nolistsep]
\item An ABD event corresponds to the class $\mathcal{C}_1$; we have 672 labeled instances.
\item No ABD event corresponds to the class $\mathcal{C}_2$; we have 533866 labeled instances.
\end{enumerate}
The target corresponding to a feature vector $\mathbf{x} \in \mathbb{R}^{30}$ is denoted by a vector $\mathbf{t}$. For the class $\mathcal{C}_1$, we use the target $\mathbf{t} = [1 \,\, 0]^{\top}$ and, for class $\mathcal{C}_2$, we use $\mathbf{t} = [0 \,\, 1]^{\top}$. The ABD-events have been manually labeled on the signals in cooperation with medical specialists. Figure \ref{fig:blackbox} shows $x_1$ and $x_2$ signals for a randomly chosen patient, and occurrence of ABD events manually labeled. A frame is labeled as the class $C_1$ if the middle of the frame is a green line. All other frames correspond to class $C_2$. 
\def\scale{.1}

\section{Methods}
\vspace{-0.2cm}
\subsection{Machine learning algorithms}
\vspace{-0.1cm}
\subsubsection{Progressive learning network (PLN)}
\vspace{-0.1cm}
In \cite{Chatterjee2017}, the PLN is shown to address pertinent technical questions on designing large size neural networks. The questions are: (a) How to determine the number of layers and the number of hidden neuron nodes in a neural network? (b) How to avoid overfitting to training data using appropriate regularization? (3) How to reduce effort in tuning of parameters in the network? The success of the PLN to address the technical questions makes it a user friendly machine learning tool. The user does not require to do pain-stalking tuning of many parameters in the PLN. Further the user can hope for a machine learning solution where chance of over-fitting is diminished. These advantages of PLN can be exploited for machine learning in medical data analytics, where medical practitioners might not have thorough technical knowledge to deal with existing neural networks. Using a mathematical property of rectifier-linear-unit (ReLU) activation function -- called progression property in \cite{Chatterjee2017} --, the training algorithm generates a self-organizing PLN. In the training algorithm, the learning is performed in layer-wise fashion. A new layer is added on an existing optimized network to grow the network further, makes it wide and deep. For learning each layer, PLN uses a convex optimization problem with appropriate regularization. The regularization helps to address the problem of overfitting. In \cite{Chatterjee2017}, the PLN was shown to provide high quality performances for many standard pattern classification and regression tasks  where it was also compared with least-squares based learning methods and extreme learning machines \cite{Huang_Extreme_learning_machine_2006}. For the ABD event detection, we are interested to see how the PLN performs when the training data is unbalanced and limited in amount.
\vspace{-0.2cm}
\subsubsection{SVM and ANN}
\vspace{-0.2cm}
Support vector machine (SVM) \cite{bishop2006} is based on a kernel approach that uses a high dimensional non-linear projection of feature vectors and then uses a linear classifier in the transformed feature space. For a binary classification task, SVM provides an optimal solution (of a suitable cost function) that provides a maximum margin between two classes that are linearly separable in a high dimensional feature space. SVM uses kernel trick and it is a maximum margin classifier. It also uses training data sparsely and is expected to be less prone to unbalanced dataset. We have used Gaussian kernels for realizing SVM in our experiments. Then we used a cross-entropy minimization based artificial neural network (ANN) \cite{bishop2006}. A feature vector $\mathbf{x}$ is an input to the ANN and then the input passes through several linear and non-linear transforms to produce the estimated class of the input feature vector. The weights of ANN are learned via a back-propagation algorithm (a gradient descent on the cross-entropy loss function).  We used a single layer feed forward ANN as we had access to a limited amount of training data. 
\vspace{-0.2cm}
\subsection{Unbalanced data and performance measure}\label{subsec:unbalanced_data}
\vspace{-0.2cm}
The size of training dataset for class $\mathcal{C}_1$ is $0.13\%$ of the size of the training dataset for class $\mathcal{C}_2$. Direct use of such an unbalanced training dataset leads to overfitting. To reduce the effect of overfitting, we endeavor to balance the training dataset by sampling feature vectors from class $\mathcal{C}_1$ and duplicating them to create a larger dataset for class $\mathcal{C}_1$. 
The training dataset is thus an artificially balanced dataset. For the testing dataset, the classification performance is assessed using the geometric mean between precision and recall, also called Fowlkes-Mallows index: $g(\mathbf{t},\hat{\mathbf{t}})=\sqrt{\frac{TP}{TP+FP} \frac{TP}{TP+FN}}$ \cite{Fowlkes1983}, where TP is the number of true positive $(\mathbf{t},\hat{\mathbf{t}})=([1 \,\, 0]^{\top},[1 \,\, 0]^{\top})$, FP is the number of false positives $(\mathbf{t},\hat{\mathbf{t}})=([0 \,\, 1]^{\top},[1 \,\, 0]^{\top})$, FN is the number of false negatives $(\mathbf{t},\hat{\mathbf{t}})=([1 \,\, 0]^{\top},[0 \,\, 1]^{\top})$; here $\hat{\mathbf{t}}$ is the output of an algorithm. For an algorithm, if the average Fowlkes-Mallows index(mean index) is high then the algorithm is good. Also if the standard deviation of the Fowlkes-Mallows index is low then the algorithm is good. Therefore, a good trade-off between high average and low standard deviation of Fowlkes-Mallows index is required for a preferred algorithm.

\section{Experimental results}
\begin{figure}[ht!]
\caption{Experimental results for ANN, SVM and PLN}
\begin{subfigure}{.49\textwidth}
\caption{Testing performance (average Fowlkes-Mallows index and standard deviation)}
\label{table:Algo_Performance2}
\def\nsys{4}
\def\nres{3}
\renewcommand{\baselinestretch}{1}
\centering
\scriptsize
\setlength{\tabcolsep}{3.5pt}
	\begin{tabular}{|c|c|c|c|}
	\hline
	Model & ANN & SVM & PLN\\ \hline
	Parameters	& 1 layer, 400 hidden nodes&Gaussian kernel&$(\lambda,\mu)=(0.1,1e4)$\\ \hline
	   & \multicolumn{\nres}{|c|}{Testing performance}		 \\ \hline
	Feature replication & $0.03 \pm 0.03$& $0.28 \pm 0.18$&$0.22 \pm 0.06$\\ \hline	
	\end{tabular}
\end{subfigure}\hfill
\begin{subfigure}{.32\textwidth}
		\centering
		\caption{Recall - Precision plane}
		\includegraphics[width=\textwidth]{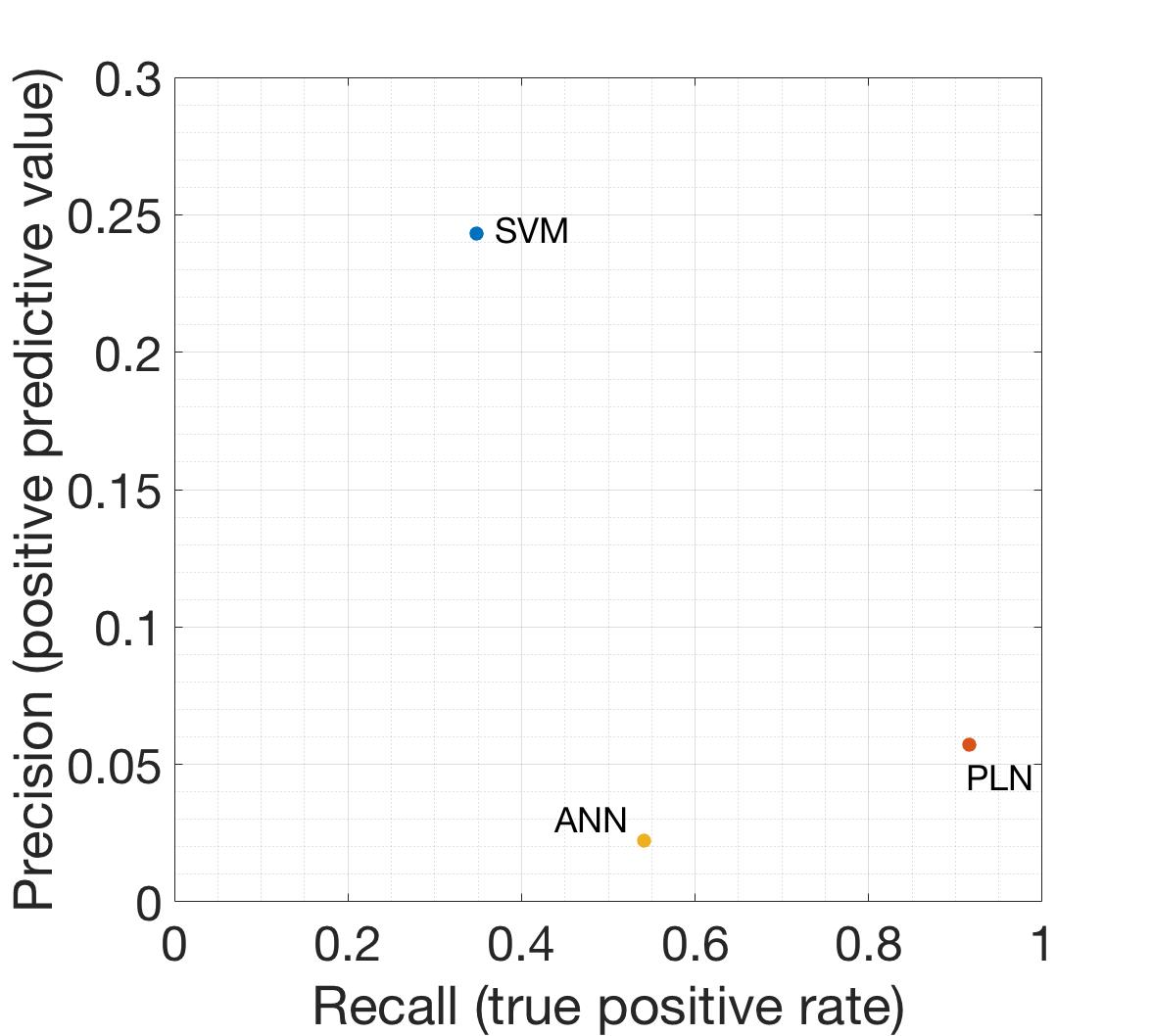}

		\label{fig:2dplan}
	\end{subfigure}
\end{figure}

\vspace{-0.3cm}
\label{sec:Experimental_results}
For a statistically relevant study with limited number of patients, we use a leave-one-out strategy allowing cross-validation for the entire data. We segregate one patient and use the data for that patient as testing dataset and the remaining 12 patients as the training dataset. Each patient is used once for testing and we report the average and standard deviation of the Fowlkes-Mallows index over the 13 trials. Table \ref{table:Algo_Performance2} shows the generalization performance of the three algorithms that we compared. In the table, we also mentioned some relevant parameters of the algorithm that we set so that the results are reproducible. Note that ANN provides poorest performance. We used a single layer feed forward network with 400 hidden neurons and sigmoid activation function. The results of ANN confirm that the detection that we are dealing with in non-trivial. The SVM provides good average performance, but shows high standard deviation. This indicates that SVM works well for several patients, but the detection fails for few patients. At the last, PLN provides a good trade-off between average value and standard deviation of the Fowlkes-Mallows index. This result is encouraging as it shows that PLN works across all 13 patients in a similar fashion.

Next, we plot average precision and average recall results in Figure \ref{fig:2dplan}. 
Note that SVM has a high precision and low recall whereas the PLN has a low precision, but high recall.
A high recall means that PLN provides a good detection quality for the ABD events (high true positives and low false negatives). At the same time, the low precision result of PLN means that it provides a large amount of false positives. This happened due to the duplication of ABD events in training data, that biased the PLN towards detecting ABD event even though there is no ABD event. This result shows that the duplication in training data to avoid overfitting is a critical task. As SVM uses training data sparsely, the effect of unbalanced data is less pronounced in SVM in the average sense, providing quasi similar values of precision and recall.
SVM triggers less false alarms (higher precision) but missed to detect ABD events more frequently than PLN (lower recall).
Overall, we find that the PLN can be viewed as the best performing algorithm among the three algorithms.

\section{Conclusions}
\vspace{-0.2cm}
For detection of ABD events, we have investigated three algorithms in this article. Two of them are widely used and one of them is recently proposed. The new progressive learning network (PLN) algorithm is found to be the best. The result for a standard artificial neural network (ANN) based on cross-entropy loss minimization is pessimistic. The support vector machine is promising, but suffers in high variability across patients. Naturally, for further improvement a direction is to exploit the technical advantages and structures of support vector machine in the framework of progressive learning network. Another important line of study is how to handle the limited as well as unbalanced data. For medical data analytics, it is natural that we will continue to deal with limited data in many situations as well as unbalanced data - more data labeled with no clinical symptoms and less data with clinical symptoms. By its nature, the machine learning system will be improved by including more patients. A better design of machine learning algorithm is also expected to contribute to a substantial improvement in detection performance. This should lead to better care of patients.

\newpage  

%


\bibliography{library}
\bibliographystyle{IEEEtran}

\end{document}